\documentclass[wcp]{jmlr}

\usepackage{booktabs}

\usepackage{graphicx}
\usepackage{float}
\usepackage{amssymb}
\usepackage{amsmath}
\usepackage{adjustbox}
\usepackage{subcaption}
\usepackage{booktabs}  
\usepackage{url,xcolor}
\usepackage{amsfonts}       
\usepackage{nicefrac}       
\usepackage{microtype}      
\usepackage{multirow}
\usepackage{amsmath}
\usepackage[misc]{ifsym}

\usepackage{makecell}
\usepackage[T1]{fontenc}
\usepackage{wrapfig}

\usepackage{algorithm}

\makeatletter
\newcommand{\AlgoResetCount}{\renewcommand{\@ResetCounterIfNeeded}{\setcounter{AlgoLine}{0}}}
\newcommand{\AlgoNoResetCount}{\renewcommand{\@ResetCounterIfNeeded}{}}
\newcounter{AlgoSavedLineCount}

\makeatother

\jmlryear{2020}
\jmlrworkshop{Accepted to the Proceedings of ACML 2020}

\title[Learning 2-opt Heuristics for the TSP via Deep Reinforcement Learning]{Learning 2-opt Heuristics for the Traveling Salesman Problem via Deep Reinforcement Learning}

  \author{\Name{Paulo Roberto de O. {da Costa}} \Email{p.r.d.oliveira.da.costa@tue.nl}\\
  \Name{Jason Rhuggenaath} \Email{j.s.rhuggenaath@tue.nl}\\
  \Name{Yingqian Zhang} \Email{yqzhang@tue.nl}\\
  \Name{Alp Akcay} \Email{a.e.akcay@tue.nl}\\
  \addr School of Industrial Engineering\\
  Eindhoven University of Technology, 5612 AZ Eindhoven, Netherlands}

\begin{document}

\maketitle

\begin{abstract}
Recent works using deep learning to solve the Traveling Salesman Problem (TSP) have focused on learning construction heuristics. Such approaches find TSP solutions of good quality but require additional procedures such as beam search and sampling to improve solutions and achieve state-of-the-art performance. However, few studies have focused on improvement heuristics, where a given solution is improved until reaching a near-optimal one. In this work, we propose to learn a local search heuristic based on 2-opt operators via deep reinforcement learning. We propose a policy gradient algorithm to learn a stochastic policy that selects 2-opt operations given a current solution. Moreover, we introduce a policy neural network that leverages a pointing attention mechanism, which unlike previous works, can be easily extended to more general $k$-opt moves. Our results show that the learned policies can improve even over random initial solutions and approach near-optimal solutions at a faster rate than previous state-of-the-art deep learning methods.
\end{abstract}

\section{Introduction}

The Traveling Salesman Problem (TSP) is a well-known combinatorial optimization problem. In the TSP, given a set of locations (nodes) in a graph, we need to find the shortest tour that visits each location exactly once and returns to the departing location. The TSP is NP-hard \citep{papadimitriou1977euclidean} even in its Euclidean formulation, i.e., nodes are points in the 2D space. Classic approaches to solve the TSP can be classified in exact and heuristic methods. The former have been extensively studied using integer linear programming \citep{applegate2006traveling} 
which are guaranteed to find an optimal solution but are often too computationally expensive to be used in practice. The latter are based on (meta)heuristics and approximate algorithms \citep{arora1998polynomial} that 
find solutions requiring less computational time, e.g., edge swaps such as $k$-opt \citep{helsgaun2009general}. Nevertheless, designed heuristics require specialized knowledge and their performances are often limited by algorithmic design decisions.

Recent works in machine learning and deep learning have focused on learning heuristics for combinatorial optimization problems \citep{bengio2018machine,lombardi2018boosting}. For the TSP, both supervised learning \citep{vinyals2015pointer,joshi2019efficient} and reinforcement learning \citep{Bello2017WorkshopT,wu2019learning,kool2018attention,deudon2018learning,khalil2017learning} methods have been proposed. The idea is that a machine learning method could potentially learn better heuristics by extracting useful information directly from data, rather than having an explicitly programmed behavior. Most approaches to the TSP have focused on learning construction heuristics, i.e., methods that can generate a solution sequentially by extending a partial tour. These methods employed sequence representations \citep{vinyals2015pointer,Bello2017WorkshopT}, graph neural networks \citep{khalil2017learning,joshi2019efficient} and attention mechanisms \citep{kool2018attention,deudon2018learning,wu2019learning} resulting in high-quality solutions. However, construction methods still require additional procedures such as beam search, classical improvement heuristics and sampling to achieve such results. This limitation hinders their applicability as it is required to revert back to handcrafted improvement heuristics and search algorithms for state-of-the-art performance.  Thus, learning improvement heuristics, i.e., 
when a solution is improved by local moves that search for better solutions, remains relevant. 
Here the idea is that if we can learn a policy to improve a solution, we can use it to get better solutions from a construction heuristic or even random solutions. Recently, a deep reinforcement learning method \citep{wu2019learning}
has been proposed for such task, achieving near-optional results using node swap and 2-opt moves. However, the architecture has its output fixed by the number of possible moves, making it less favorable to expand to more general $k$-opt moves. 


In this work, we propose a deep reinforcement learning algorithm trained via Policy Gradient to learn improvement heuristics based on 2-opt moves. Our architecture is based on a pointer attention mechanism \citep{vinyals2015pointer} that outputs nodes sequentially for action selection. We introduce a reinforcement learning formulation to learn a stochastic policy of the next promising solutions, incorporating the search's history information by keeping track of the current best-visited solution. Our results show that we can learn policies for the Euclidean TSP that achieve near-optimal solutions even when starting from solutions of poor quality. Moreover, our approach can achieve better results than previous deep learning methods based on construction \citep{vinyals2015pointer,joshi2019efficient,kool2018attention,deudon2018learning,khalil2017learning,Bello2017WorkshopT} and improvement \citep{wu2019learning} heuristics. Compared to \cite{wu2019learning}, our method can be easily adapted to general $k$-opt and it is more sample efficient. In addition, policies trained on small instances can be reused on larger instances of the TSP. Lastly, our method outperforms other effective heuristics such as Google's OR-Tools \citep{ortools} and are close to optimal solutions.

\section{Related Work}


In machine learning, early works for the TSP have focused on Hopfield networks \citep{hopfield1985neural} and deformable template models \citep{angeniol1988self}. However, the performance of these approaches has not been on par with classical heuristics \citep{la2012comparison}. Recent deep learning methods have achieved high performance learning construction heuristics for the TSP. Pointer Networks (PtrNet) \citep{vinyals2015pointer} learned a sequence model coupled with an attention mechanism trained to output TSP tours using solutions generated by Concorde \citep{applegate2006traveling}. 
In \cite{Bello2017WorkshopT}, the PtrNet was further extended to learn without supervision using Policy Gradient, trained to output a distribution over node permutations. 
Other approaches encoded instances via graph neural networks. A \emph{structure2vec} (S2V) \citep{khalil2017learning} model was trained to output the ordering of partial tours using Deep Q-Learning (DQN). Later, graph attention was employed to a hybrid approach using 2-opt local search on top of tours trained via Policy Gradient \citep{deudon2018learning}. Graph attention was extended in \cite{kool2018attention} using REINFORCE \citep{williams1992simple} with a greedy rollout baseline, resulting in lower optimality gaps. Recently, the supervised approach was revisited using Graph Convolution Networks (GCN) \citep{joshi2019efficient} learning probabilities of edges occurring in a TSP tour. It achieved state-of-the-art results up to 100 nodes whilst also combining with search heuristics.

It is important to previous methods to have additional procedures such as beam search, classical improvement heuristics and sampling to achieve good solutions. However, little attention has been posed on learning such policies that search for improved solutions. A recent approach, based on the transformer architecture \citep{wu2019learning}, employed a Graph Attention Network (GAT) \citep{velickovic2018graph} coupled with 2-opt and node swap operations. The limitations of this approach are related to the fixed output embeddings. These are vectors defined by the squared number of nodes, which makes expanding to general $k$-opt harder. Moreover, at inference it requires more samples than construction methods to achieve similar results. In contrast, we encode edge information using graph convolutions and use classical sequence encoding to learn tour representations. We decode these representations via a pointing attention mechanism to learn a stochastic policy of the action selection task. Our approach resembles classical 2-opt heuristics \citep{hansen2006first} and can outperform previous deep learning methods in solution quality and sample efficiency.

\section{Background}

\subsection{Travelling Salesman Problem}

We focus on the 2D Euclidean TSP. Given an input graph, represented as a sequence of $n$ locations in a two dimensional space $X = \{x_i\}^n_{i=1}$, where $x_i \in [0, 1]^2$, we are concerned with
finding a permutation of the nodes, i.e. a tour $S = (s_1, \dots, s_n)$, that visits each node once (except the starting node) and has the minimum total length (cost). We define the cost of a tour as the sum of the distances (edges) between consecutive nodes in $S$ as $ L(S)= \left\|x_{s_n}-x_{s_1} \right\|_2 + \sum_{i=1}^{n-1}\left\| x_{s_i}-x_{s_{i+1}}\right\|_2$, 
where $ \left\|\cdot\right\|_2$ denotes the $\ell_2$ norm.

\subsection{$k$-opt Heuristic for the TSP}

\begin{figure}[ht!]
    \centering
    \begin{center}
    \includegraphics[width=0.4\textwidth]{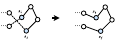}    
    \end{center}
     \caption{TSP solution before a 2-opt move (left), and after a 2-opt move (right). Replaced edges are represented in dashed lines. Note that the sequence $s_i, \dots, s_j$ is inverted.}
      \label{fig:operator}

\end{figure}

Improvement heuristics enhance feasible solutions through a search procedure. A procedure starts at an initial solution $S_0$ and replaces a previous solution $S_t$ by a better solution $S_{t+1}$. Local search methods such as the effective Lin-Kernighan-Helsgaun (LKH) \citep{helsgaun2009general} heuristic perform well for the TSP. The procedure searches for $k$ edge swaps ($k$-opt moves) that will be replaced by new edges resulting in a shorter tour. A simpler version \citep{lin1973effective} considers 2-opt (Figure \ref{fig:operator}) and 3-opt moves alternatives as these balance solution quality and the $O(n^k)$ complexity of the moves. Moreover, sequential pairwise operators such as $k$-opt moves can be decomposed in simpler $l$-opt ones, where $l<k$. For instance, sequential 3-opt operations can be decomposed into one, two or three 2-opt operations \citep{helsgaun2009general}. However, in local search algorithms, the quality of the initial solution usually affects the quality of the final solution, i.e. local search methods can easily get stuck in local optima \citep{hansen2006first}.

To avoid local optima, different metaheuristics have been proposed including Simulated Annealing and Tabu Search. These work by accepting worse solutions to allow more exploration of the search space. 
In general, this strategy leads to better solution quality. However, metaheuristics still require expert knowledge and may have sub-optimal rules in their design. To tackle this limitation, we propose to combine machine learning and 2-opt operators to learn a stochastic policy to improve TSP solutions sequentially. A stochastic policy resembles a metaheuristic, sampling solutions in the e neighborhood of a given solution, 
potentially avoiding local minima. Our policy iterates over feasible solutions and the minimum cost solution is returned at the end. The main idea of our method is that by taking future improvements into account can potentially result it better policies than greedy heuristics. 

\section{Reinforcement Learning Formulation}

Our formulation considers solving the TSP via 2-opt as a Markov Decision Process (MDP), detailed below. 
In our MDP, a given state $\Bar{S}$ is composed of a tuple of the current solution (tour) $S$ and the lowest-cost solution $S^\prime$ seen in the search.
The proposed \emph{neural architecture} (Section \ref{sec:PGN}) approximates the stochastic policy $\pi_{\theta}(A|\Bar{S})$, 
 where $\theta$ represents trainable parameters.
Each $A = (a_1, a_2)$ corresponds to a 2-opt move where $a_1,a_2$ are node indices. Our architecture also contains a \emph{value} network that outputs value estimates $V_\phi(\bar{S})$, with $\phi$ as learnable parameters. We assume  TSP samples drawn from the same distribution 
and use Policy Gradient to optimize the parameters of the policy and value networks (Section \ref{sec:PGO}).



\paragraph{States} A state $\Bar{S}$ is composed of a tuple $\Bar{S} = (S, S^\prime)$, where $S$ and $S^\prime$ are the current and lowest-cost solution seen in the search, respectively. That is, given a search trajectory at time $t$ and solution $S$,
$S_t = S$ and $S^{\prime}_t = S^\prime = \text{arg\,min}_{S_{\tilde{t}} \in \{ S_0, \ldots, S_t\}}  L(S_{\tilde{t}})$.

\paragraph{Actions} 
We model actions as tuples $ A = (a_1, a_2)$ where $a_1,a_2 \in \{1,\dots,n\}$, $a_2 > a_1$ correspond to index positions of solution $S = (s_1,\dots,s_n)$.

\paragraph{Transitions} Given $A=(i,j)$ transitioning to the next state defines a deterministic change to solution $\hat{S} = (\dots,s_i, \dots, s_j, \dots)$, resulting in a new solution $S  = (\dots,s_{i-1},s_j, \dots, s_i, s_{j+1} \dots) $ and state $\Bar{S} = (S, S^\prime)$. That is, selecting $i$ and $j$ in $\hat{S}$ implies breaking edges at positions $(i-1, i)$ and $(j, j+1)$, inserting edges $(i-1, j)$ and $(i, j+1)$ and inverting the order of nodes between $i$ and $j$.

\paragraph{Rewards} Similar to \citep{wu2019learning}, we attribute rewards to actions that can improve upon the current best-found solution, i.e., 
$R_{t} = L(S^{\prime}_t) - L(S^{\prime}_{t+1})$.

\paragraph{Environment} Our environment runs for $\mathbb{T}$ steps. For each run, we define episodes of length $T \leq \mathbb{T}$, after which a new episode starts from the last state in the previous episode. This ensures access to poor quality solutions at $t=0$, and high quality solutions as $t$ grows. In our experiments, treating the environment as continuous and bootstrapping \citep{mnih2016asynchronous} resulted in lower quality policies under the same conditions. 

\paragraph{Returns} Our objective is to maximize the expected return $G_t$, which is the cumulative reward starting at time step $t$ and finishing at $T$ at which point no future rewards are available, i.e., $G_t =\sum_{t^\prime=t}^{T-1} \gamma^{t^\prime-t}  R_{t^\prime} $ where $\gamma \in (0, 1]$ is a discount factor.

\section{Policy Gradient Neural Architecture}
\label{sec:PGN}
\begin{figure*}
    \centering
    \includegraphics[width=0.95\textwidth]{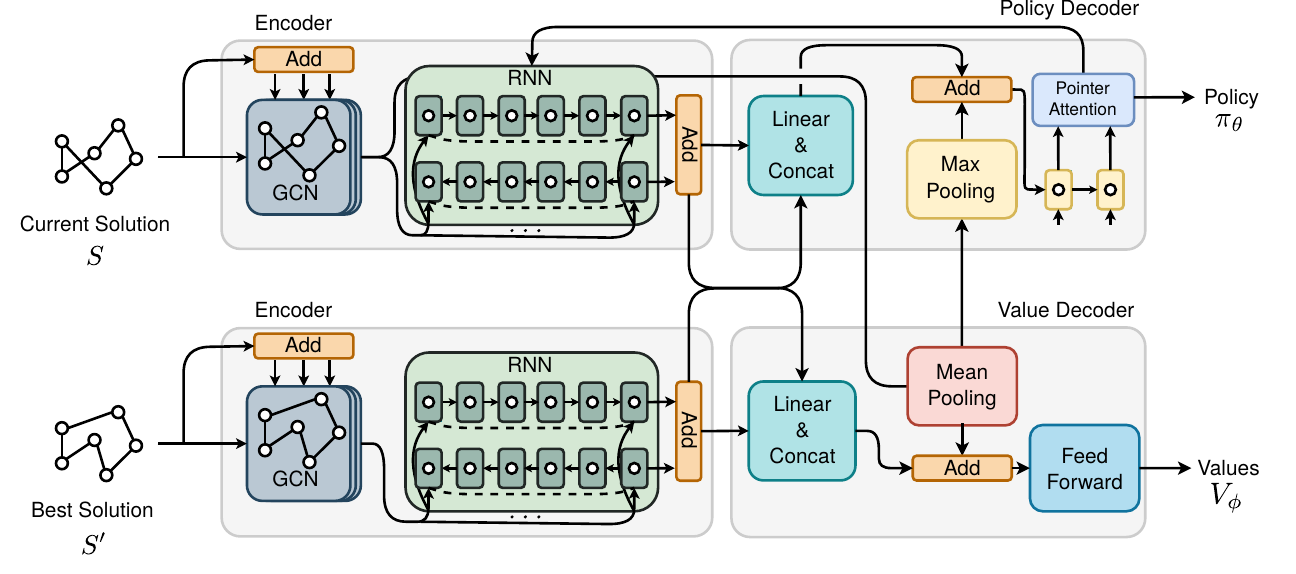}
    \caption{In the architecture, a state $\Bar{S} = (S, S^\prime) $ is passed to a dual encoder where graph and sequence information are extracted.  
    A policy decoder takes encoded inputs to query node indices and output actions. A value decoder takes encoded inputs and outputs state values.}
    \label{fig:nn}
\end{figure*}

Our neural network, based on an encoder-decoder architecture is depicted in Figure \ref{fig:nn}. Two encoder units map each component of $\Bar{S} = (S, S^\prime)$ independently. Each unit reads inputs $X = (x_1, \dots, x_n)$, where $x_i$ are node coordinates of node $s_i$ in $S$ and $S^\prime$ . The encoder then learns representations that embed both graph topology and node ordering. Given these representations, the \emph{policy} decoder samples action indices $a_1, \dots, a_k$ sequentially, where $k=2$ for 2-opt. The \emph{value} decoder operates on the same encoder outputs but outputs real-valued estimates of state values. We detail the components of the network in the following sections.

\subsection{Encoder}

The purpose of our encoder is to obtain a representation for each node in the input graph given its topological structure and its position in a given solution. To accomplish this objective, we incorporate elements from Graph Convolution Networks (GCN) \citep{kipf2016semi} and sequence embedding via Recurrent Neural Networks (RNN) \citep{hochreiter1997long}. Furthermore, we use edge information to build a more informative encoding of the TSP graph. 

\paragraph{Embedding Layer}

We input two dimensional coordinates $x_i \in {[0,1]}^2$, $ \forall i \in 1,\ldots, n $, which are embedded to $d$-dimensional features as
$x^0_i = W_x  x_i + b_x\,,$ 
where $W_x\in \mathbb{R}^{d \times 2}$, $b_x\in \mathbb{R}^{d}$. We use as input the Euclidean distances $e_{i,j}$ between coordinates $x_i$ and $x_j$ to add edge information and weigh the node feature matrix. To avoid scaling the inputs to different magnitudes we adopt symmetric normalization \citep{kipf2016semi} as $  \tilde{e}_{i,j}= \frac{e_{i,j}}{\sqrt{\sum_{j=1}^n e_{i,j} \sum_{i=1}^n e_{i,j}}}\,.$
%
%
Then the normalized edges are used in combination with GCN layers to create richer node representations using its neighboring topology. 

\paragraph{Graph Convolutional Layers}

In the GCN layers, we denote as $x_i^{\ell}$ the node feature vector at GCN layer $\ell$ associated with node $i$. We define the node feature at the subsequent layer combining features from nodes in the neighborhood $\mathcal{N}(i)$ of node $i$ as
\begin{equation}
    \label{eq:4}
    x_i^{\ell+1} = x_i^{\ell} + \text{ReLU} \Big(\sum\nolimits_{j\in \mathcal{N}(i)} \tilde{e}_{i,j}  (W_g^\ell x_j^{\ell}+b_g^\ell) \Big), 
\end{equation}%
where $W^{\ell}_g \in \mathbb{R}^{d \times d} $, $b^{\ell}_g \in \mathbb{R}^{d}$, $\text{ReLU}$ is the Rectified Linear Unit and  $\mathcal{N}(i)$ corresponds to the remaining $n-1$ nodes of a complete TSP network. At the input to these layers, we have $\ell=0$ and after $\mathbb{L}$ layers we arrive at representations $z_i = x_i^{\mathbb{L}}$ leveraging node features with the additional edge feature representation.

\paragraph{Sequence Embedding Layers}

Next, we use node embeddings $z_i$ to learn a sequence representation of the input and encode a tour. Due to symmetry, a tour from nodes $(1,\ldots,n)$ has the same cost as the tour $(n,\ldots,1)$. Therefore, we read the sequence in both orders to explicitly encode the symmetry of a solution. To accomplish this objective, we employ two Long Short-Term Memory (LSTM) as our RNN functions, computed using hidden vectors from the previous node in the tour and the current node embedding resulting in
\begin{equation}
    \label{eq:5}
    (h^{\rightarrow}_i, c^{\rightarrow}_i) = \text{RNN}(z_i^{\rightarrow}, (h_{i-1}^{\rightarrow},c_{i-1}^{\rightarrow}) ), \quad  i \in (1,\ldots,n)
\end{equation}
\begin{equation}
    \label{eq:7}
    (h^{\leftarrow}_i, c^{\leftarrow}_i) = \text{RNN}(z_i^{\leftarrow}, (h_{i+1}^{\leftarrow},c_{i+1}^{\leftarrow}) ),  \quad  i \in (n,\ldots,1)
\end{equation}
where in (\ref{eq:5}) a forward RNN goes over the embedded nodes from left to right, in (\ref{eq:7}) a backward RNN goes over the nodes from right to left and $h_i, c_i \in \mathbb{R}^d$ are hidden vectors. 

Our representation reconnects back to the first node in the tour ensuring we construct a sequential representation of the complete tour, i.e. $(h^{\rightarrow}_0, c^{\rightarrow}_0) = \text{RNN}(z_n, 0)$ and  $(h^{\leftarrow}_{n+1}, c^{\leftarrow}_{n+1}) = \text{RNN}(z_1, 0)$.
Afterwards, we combine forward and backward representations to form unique node representations in a tour  as $o_{i} = \text{tanh}((W_fh^{\rightarrow}_i + b_f) + (W_bh^{\leftarrow}_i + b_b))$, and a tour representation $h_n = h^{\rightarrow}_n + h^{\leftarrow}_n$, where $h_i, o_i \in \mathbb{R}^d$, $W_f, W_b \in \mathbb{R}^{d \times d}$ and $b_f, b_b \in \mathbb{R}^{d}$.

\paragraph{Dual Encoding} In our formulation, a state $\Bar{S} = (S, S^\prime) $ is represented as a tuple of the current solution $S$ and the best solution seen so far $S^\prime$. For that reason, 
we encode both $S$ and $S^\prime$ using independent encoding layers (Figure \ref{fig:nn}). Thus, we abuse notation and define a sequential representation of $S^\prime$ after going through encoding layers as $h^{\prime}_n \in \mathbb{R}^d$.

\subsection{Policy Decoder}

We aim to learn the parameters of a stochastic policy $\pi_\theta (A| \Bar{S})$ that given a state $\Bar{S}$, assigns high probabilities to moves that reduce the cost of a tour. Following \citep{Bello2017WorkshopT}, our architecture uses the chain rule to factorize the probability of a $k$-opt move as
\begin{equation}
  \pi_{\theta}(A|\Bar{S}) = \prod_{i=1}^{k} p_\theta\left(a_i | a_{<i}\,,
  \Bar{S}\right),
  \label{eqn:prob}
\end{equation}
and then uses individual softmax functions to represent each term on the RHS of \eqref{eqn:prob}, where $a_i$ corresponds to node positions in a tour, $a_{<i}$ represents previously sampled nodes and $k=2$. 
At each output step $i$, we map the tour embedding vectors to the following \emph{query} vector
\begin{equation}
  q_i = \tanh \Big( (W_q q_{i-1} + b_q) + (W_o o_{i-1} + b_o)\Big),
  \label{eqn:query}
\end{equation}
where $W_q, W_o \in \mathbb{R}^{d \times d}$, $b_q, b_o \in \mathbb{R}^{d \times d}$  are learnable parameters and $o_0 \in \mathbb{R}^{d}$ is a fixed parameter initialized from a uniform distribution $\mathcal{U}(\frac{-1}{\sqrt{d}}, \frac{1}{\sqrt{d}})$. Our initial query vector $q_0$ receives the tour representation from $S$ and $S^\prime$ as $h_{\Bar{s}}  =  W_s h_n + b_s \| W_{s^{\prime}} h^\prime_n + b_{s^\prime}$ and a \emph{max pooling} graph representation $z_g = \max( z_1,\ldots, z_n )$ from $S$ to form $ q_0  =  h_{\Bar{s}} + z_g,$
%
where learnable parameters $W_s, W_{s^\prime} \in \mathbb{R}^{\frac{d}{2} \times d}$, $b_s, b_{s^\prime} \in \mathbb{R}^{\frac{d}{2}}$ and $\cdot\|\cdot$ represents the concatenation operation. 
Our query vectors $q_i$ interact with a set of $n$ vectors to define a pointing distribution over the action space. As soon as the first node is sampled, the query vector updates its inputs with the previously sampled node using its sequential representation to select the subsequent nodes.

\paragraph{Pointing Mechanism}

We use a pointing mechanism to predict a distribution over node outputs given encoded actions (nodes) and a state representation (query vector). Our pointing mechanism is parameterized by two learned attention matrices $K \in \mathbb{R}^{d \times d}$ and $Q \in \mathbb{R}^{d \times d}$ and vector $v \in \mathbb{R}^{d}$ as 
\begin{equation}
  u^i_j=\begin{cases}
    v^T \tanh(K o_j + Qq_i), & \text{if $j > a_{i-1}$ }\\
    - \infty , & \text{otherwise}\,,
  \end{cases}
\end{equation}
where $    p_\theta\left(a_i \mid a_{<i},
  \Bar{S}\right) = \text{softmax}(C\tanh(u^i))$
%
predicts a distribution over $n$ actions, given
a query vector $q_i$ with $u^i \in \mathbb{R}^{n}$. We mask probabilities of nodes prior to the current $a_i$ as we only consider choices of nodes in which $a_i > a_{i-1}$ due to symmetry. This ensures a smaller action space for our model, i.e. $n(n-1)/2$ possible feasible permutations of the input. We clip logits in $[-C, +C]$ \citep{Bello2017WorkshopT}, where $C \in \mathbb{R}$ is a parameter to control the entropy of $u^i$. 

\subsection{Value Decoder}

Similar to the policy decoder, our value decoder works by reading tour representations from $S$ and $S^\prime$  and a graph representation from $S$. That is, given embeddings $Z$ the value decoder works by reading the outputs $z_i$ for each node in the tour and the sequence hidden vectors $h_n, h^\prime_n$ to estimate the value of a state as
\begin{equation}
\label{eq:mean_pooling}
     V_\phi(\Bar{S}) = W_r~\text{ReLU}\Big(W_z\Big(\frac{1}{n}\sum_{i=1}^n z_i + h_v\Big)+b_z\Big) + b_r\,,
\end{equation}
with $ h_{v}  =  W_v h_n + b_v\| W_{v^{\prime}} h^\prime_n + b_{v^{\prime}}$.
Where $W_z \in \mathbb{R}^{d \times d}$, $W_r \in \mathbb{R}^{1 \times d}$, $b_z \in \mathbb{R}^{d}$, $b_r \in \mathbb{R}$ are learned parameters that map the state representation to a real valued output and $W_v, W_{v^{\prime}}  \in \mathbb{R}^{\frac{d}{2} \times d}$, $b_v, b_{v^{\prime}}  \in \mathbb{R}^{\frac{d}{2}}$ map the tours to a combined value representation. 
We use a \emph{mean pooling} operation in (\ref{eq:mean_pooling}) to combine node representations $z_i$ in a single graph representation. This vector is then combined with the tour representation $h_v$ to estimate current state values. 
\section{Policy Gradient Optimization}
\label{sec:PGO}
In our formulation, we maximize the expected rewards given a state $\Bar{S}$ defined as $J(\theta \mid \Bar{S}) = \mathbb{E}_{\pi_\theta}[G_t\mid \Bar{S}]\,$.
%
Thus, during training, we define the total objective over a distribution $\mathcal{S}$ of uniformly distributed TSP graphs (solutions) in $[0, 1]^2$ as $J(\theta) = \mathbb{E}_{\Bar{S} \sim \mathcal{S}}[J(\theta \mid \Bar{S})].
    \label{eq:exp_JS}$
To optimize our policy we resort to the Policy Gradient learning rule, which provides an unbiased gradient estimate w.r.t. the model’s parameters $\theta$. 
During training, we draw $B$ i.i.d. graphs and approximate the gradient of 
(\ref{eq:exp_JS}), indexed at $t=0$ as 
\begin{equation}
    \nabla_\theta  J(\theta) \approx \frac{1}{B}\frac{1}{T}\Big[\sum_{b=1}^{B} \sum_{t=0}^{T-1} \nabla_\theta \log \pi_\theta (A^b_t \mid \Bar{S}^b_t) (G^b_t - V_\phi(\Bar{S}^b_t))\Big]
    \label{eq:polgrad}
\end{equation}
where the advantage function is defined as $ \mathcal{A}^b_t = G^b_t - V_\phi(\Bar{S}^b_{t}) $.
To avoid premature convergence to a sub-optimal policy \citep{mnih2016asynchronous}, we add an entropy bonus
\begin{equation}
    H(\theta) = \frac{1}{B} \sum_{b=1}^{B} \sum_{t=0}^{T-1} H (\pi_\theta (\cdot \mid \Bar{S}^b_t))\,,
    \label{eq:entropy}
\end{equation}
with $ H (\pi_\theta (\cdot \mid \Bar{S}^b_t)) = -\mathbb{E}_{\pi_\theta}[ \log \pi_\theta (\cdot \mid \Bar{S}^b_t)] $, and similarly to (\ref{eq:polgrad}) we normalize values in (\ref{eq:entropy}) dividing by $k$.
Moreover, we increase the length of an episode after a number of epochs, i.e. at epoch $e$, $T$ is replaced by $T_e$. The value network is trained on a mean squared error objective between its predictions and Monte Carlo estimates of the returns, formulated as an additional objective
\begin{equation}
    \mathcal{L}(\phi) = \frac{1}{B}\frac{1}{T}\Big[\sum_{b=1}^{B} \sum_{t=0}^{T-1}  \left\| G^b_t - V_\phi(\Bar{S}^b_t))\right\|^2_2\Big]\,.
    \label{eq:mse}
\end{equation}
Afterwards, we combine the previous objectives and perform gradient updates via Adaptive Moment Estimation (ADAM) \citep{kingma2015adam}, with $\beta_H ,\beta_V$ representing weights of (\ref{eq:entropy}) and (\ref{eq:mse}), respectively. Our model is close to REINFORCE \citep{williams1992simple} but with periodic episode length updates (truncation), and to Advantage Actor-Critic (A2C) \citep{mnih2016asynchronous} bootstrapping only from terminal states. In our case, this is beneficial as at the start the agent learns how to behave over small episodes for easier credit assignment, later tweaking its policy over larger horizons. The complete algorithm is depicted in Algorithm \ref{alg:training}.


\IncMargin{1.5em}
\begin{algorithm*}[ht]
\SetAlgoNoEnd
\LinesNumbered
\SetAlgoLined
    \caption{Policy Gradient Training} \label{alg:training}
    
    \KwIn{Policy network $\pi_\theta$, critic network $V_\phi$, number of epochs $\mathit{E}$, number of batches $\mathbf{N}_B$, batch size $B$, step limit $\mathbb{T}$, length of episodes $T_e$, learning rate $\lambda$}
    \BlankLine
    Initialize policy and critic parameters $\theta$ and $\phi$\;
    
    \For{ $e$ = 1, $\dots$, $\mathit{E}$}{
          $T$ $\leftarrow T_e$
          
          \For{$\mathbf{n}$ = 1, $\dots$, $\mathbf{N}_B$}{
              $t\leftarrow 0$\;
              
              Initialize random $\Bar{S}^b_0$, $\; \forall b \in \{1,\ldots,B\}$\;
              
              \While{$t< \mathbb{T}$}{
                    $t^{\prime} \leftarrow t$\;
                    
                    \While{$t-t^{\prime}< T$ }{    
                          $A^b_t   \sim\ \pi_{\theta}(.|\Bar{S}^b_t)$,    $\; \forall b \in \{1,\ldots,B\}$ \;
                          
                          Take actions $A^b_t$, observe $\Bar{S}^b_{t+1}, R^b_{t} $, $\; \forall b \in \{1,\ldots,B\}$\;
                          
                          $\Bar{S}^b_t \leftarrow  \Bar{S}^b_{t+1}$,  $\; \forall b \in \{1,\ldots,B\}$\;
                          
                          $t\leftarrow t+1$\; 
                    }
                  \For{$\mathit{i} \in \{t^\prime, \ldots, t-1\}$}{
                       $G^b_i \leftarrow \sum\limits_{\tilde{t}=i}^{t^\prime + T-1}  \gamma^{\tilde{t}-t^\prime} R^b_{\tilde{t}}  $ ,  $\; \forall b \in \{1,\ldots,B\}$\;
                   }    
                      $ g_\theta \leftarrow  \frac{1}{Bk} \Big[ \frac{1}{T} \sum\limits_{b=1}^{B} \sum\limits_{i=t^\prime}^{t-1} \nabla_\theta \log \pi_\theta (A^b_i \mid \Bar{S}^b_i) \mathcal{A}^b_i  + \beta_H  \nabla_\theta H (\pi_\theta (\cdot \mid \Bar{S}^b_i)) \Big]$\;
                      
                    $ g_\phi \leftarrow  \frac{1}{BT} \Big[ \beta_V \sum\limits_{b=1}^{B} \sum\limits_{i=t^\prime}^{t-1} \nabla_\phi \left\| G^b_t - V_\phi(\Bar{S}^b_i))\right\|^2_2 \Big]$\;
                      
                   $\theta, \phi  \leftarrow \text{ADAM}(\lambda, -g_\theta, g_\phi )$\;
             }
         }
    }
\end{algorithm*}
\DecMargin{1.5em}
\section{Experiments and Results}

We conduct extensive experiments to investigate the performance of our proposed method. We consider three benchmark tasks, Euclidean TSP with 20, 50 and 100 nodes, TSP20, TSP50 and TSP100 respectively. For all tasks, node coordinates are drawn uniformly at random in the unit square $[0, 1]^2$ during training. For validation, a fixed set of TSP instances with their respective optimal solutions is used for hyperparameter optimization. For a fair comparison, we use the \emph{same} test dataset as reported in \cite{kool2018attention, joshi2019efficient} containing 10,000 instances for each TSP size. Thus, previous results reported in \cite{kool2018attention} are comparable to ours in terms of solution quality (optimality gap). Results from \cite{wu2019learning} are not measured in the same data but use the same data generation process. Since at the time of submission no implementation is available, we report the optimality gaps reported in the original paper. 
Moreover, we report running times reported in \cite{kool2018attention, joshi2019efficient, wu2019learning}. Since time can vary due to implementations and hardware, we rerun the method of \cite{kool2018attention} in our hardware. Due to provided supervised samples, the method of \cite{joshi2019efficient} is not ideal for combinatorial problems. Thus, we compare our results in more detail to \cite{kool2018attention} (running time and solution quality) and \cite{wu2019learning} (solution quality and sample efficiency).

\subsection{Experimental Settings}

All our experiments use a similar set of hyperparameters. We use a batch size $B=512$ for TSP20 and TSP50 and $B=256$ for TSP100 due to GPU memory. For this reason, we generate 10 random mini-batches for TSP20 and TSP50 and 20 mini-batches for TSP100 in each epoch. TSP20 trains for 200 epochs as convergence is faster for smaller problems, whereas TSP50 and TSP100 train for 300 epochs. We use the same $\gamma = 0.99$, $\ell_2$ penalty of $1 \times 10^{-5}$ and learning rate $\lambda = 0.001$, $\lambda$ decaying by $0.98$ at each epoch. Loss weights are $\beta_V = 0.5$, $\beta_H = 0.0045$ for TSP20 and TSP50, $\beta_H = 0.0018$ for TSP100. $\beta_H$ decays by 0.9 after every epoch for stable convergence. In all tasks, 
$d=128$, $\mathbb{L}=3$ and we employ one RNN block. The update in episode lengths are $T_1 = 8, T_{100} = 10 , T_{150} = 20 $ for TSP 20; $T_1 = 8, T_{100} = 10 , T_{200} = 20$ for TSP50; and $T_1 = 4, T_{100} = 8 , T_{200} = 10$ for TSP100. $C=10$ is used during training and testing. $v$ is initialized as $\mathcal{U}(\frac{-1}{\sqrt{d}}, \frac{1}{\sqrt{d}})$ and remaining parameters are initialized according to PyTorch's default parameters.

We train on a RTX 2080Ti GPU, generating random feasible initial solutions on the fly at each epoch. Each epoch takes an average time of 2m 01s, 3m 05s and 7m 16s for TSP20, TSP50 and TSP100, respectively. We clip rewards to 1 to favor non-greedy actions and stabilize learning. Due to GPU memory, we employ mixed precision training \citep{jia2018highly} for TSP50 and TSP100. For comparison with \cite{wu2019learning}, we train for a maximum step limit of 200. Note that our method is more sample efficient than the proposed in \cite{wu2019learning}, using 50\% and 75\% of the total samples for TSP20 and TSP50/100 during training. During testing, we run our policy for 500, 1,000 and 2,000 steps to compare to previous works. Our implementation is available online \footnote{\url{https://github.com/paulorocosta/learning-2opt-drl}}.

\subsection{Experimental Results and Analysis}
\begin{figure}[!tbp]
  \centering
  \begin{minipage}[b]{0.308\textwidth}
 
    \includegraphics[width=\textwidth]{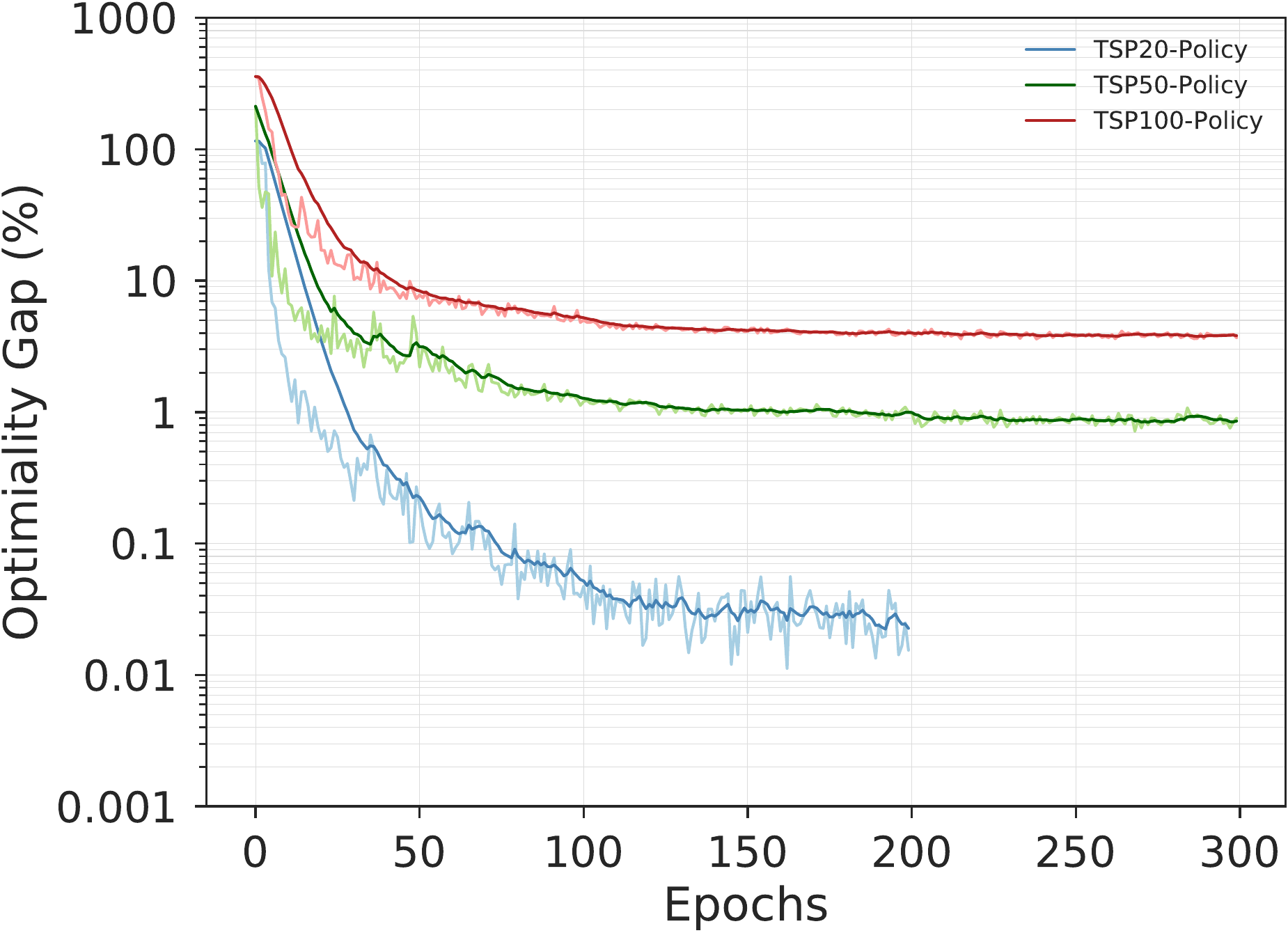}
    \caption{Optimality gaps on 256 validation instances for 200 steps over training epochs.}
    \label{fig:training}
  \end{minipage}
  \hfill
  \begin{minipage}[b]{0.308\textwidth}
    \includegraphics[width=\textwidth]{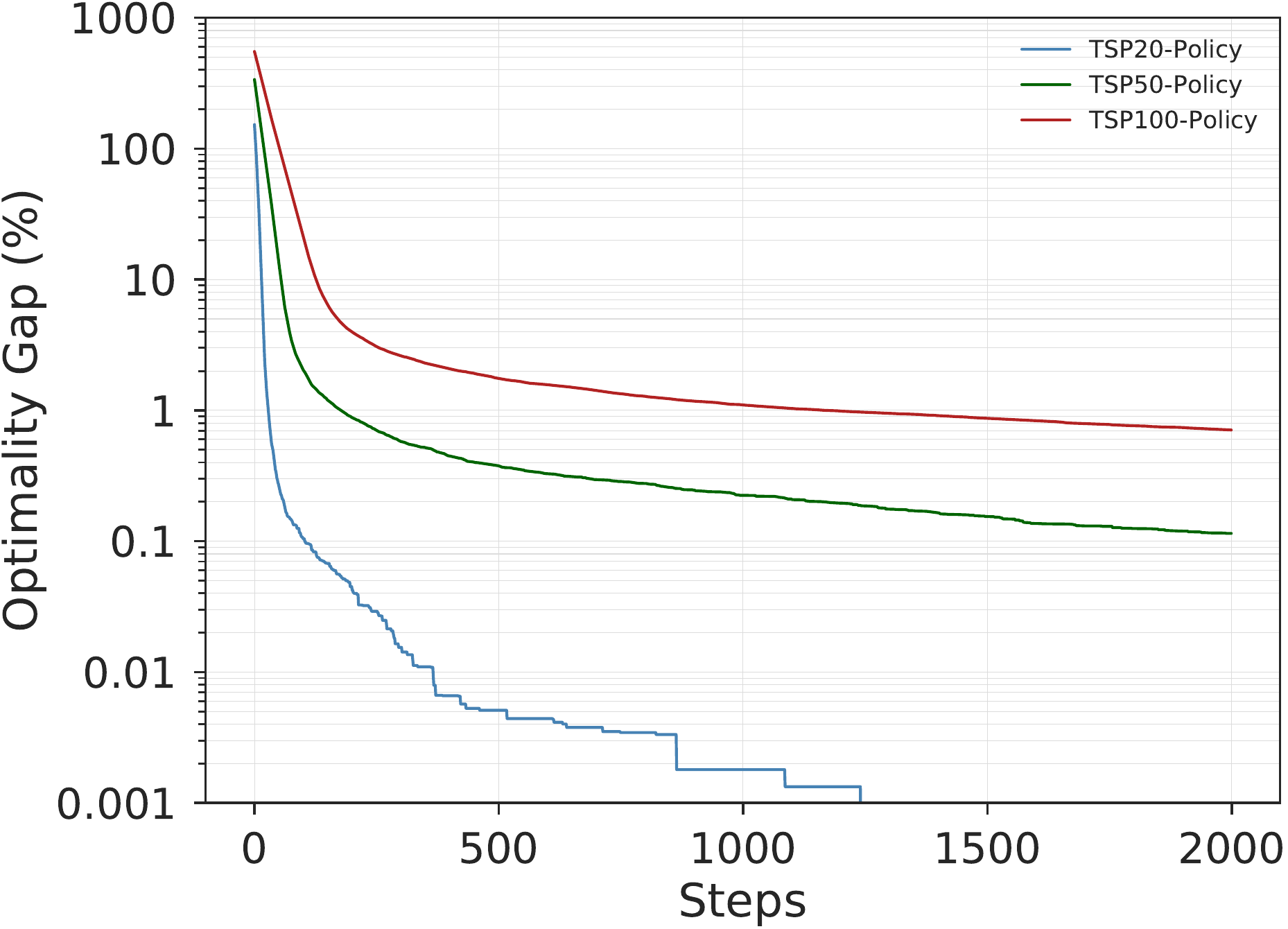}
    \caption{Optimality gaps of best found tours on 512 testing instances over 2,000 sampling steps.}
    \label{fig:testing}
  \end{minipage}
  \hfill
    \begin{minipage}[b]{0.301\textwidth}
    \includegraphics[width=\textwidth]{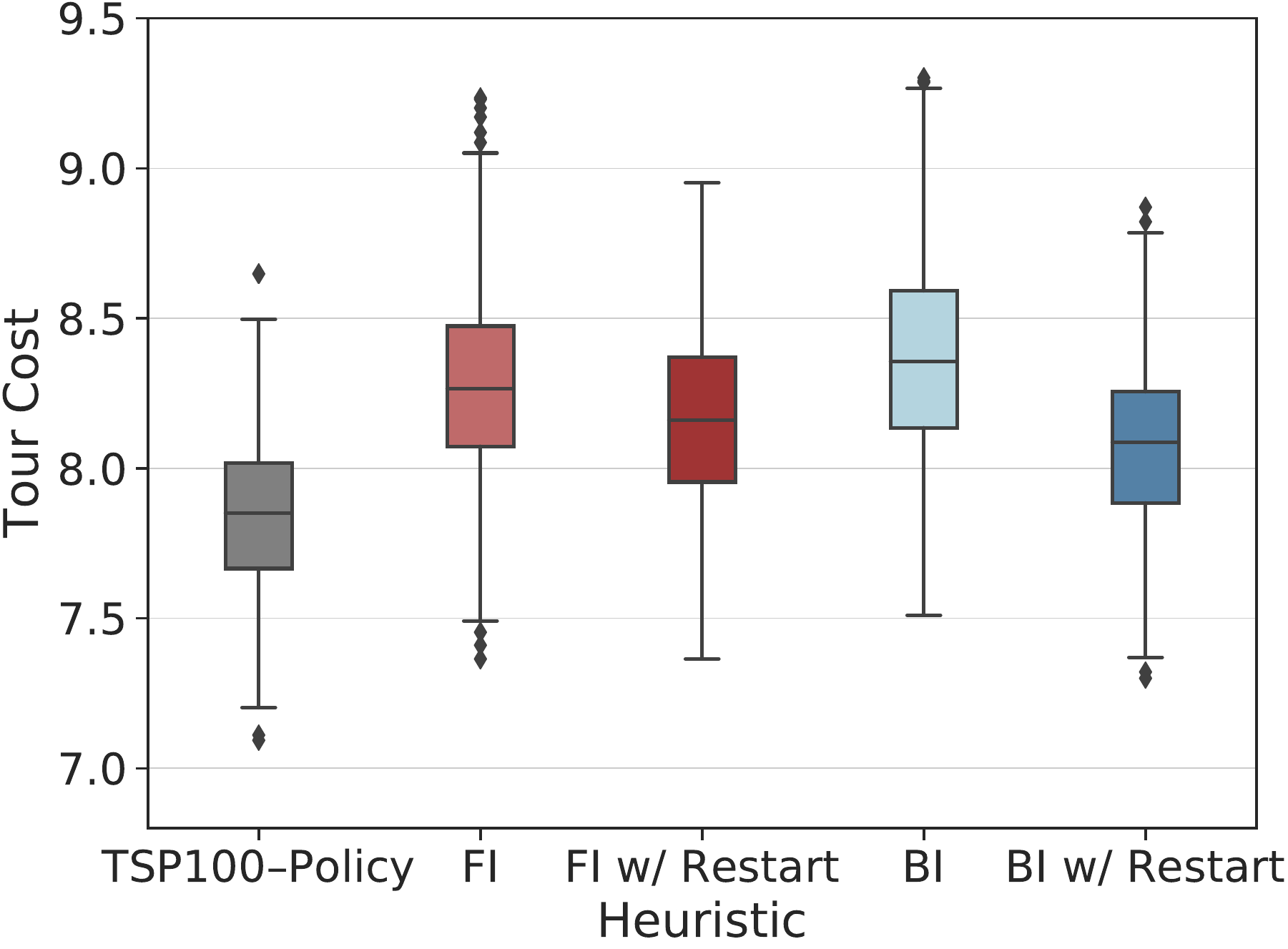}
    \caption{Tour costs of learned, FI and BI heuristics with restarts on TSP100 instances after 1,000 steps.}
    \label{fig:boxplot}
  \end{minipage}
\end{figure}

We learn policies for TSP20, TSP50 and TSP100, and depict the optimality gap and its exponential moving average in the log scale in Figure \ref{fig:training}. In the figure, the optimality gap is averaged over 256 validation instances and 200 steps (same as training). The results show that we can learn effective policies that decrease the optimality gap over the training epochs. We also point out that increasing the episode length improved validation performance as we consider longer planning horizons in (\ref{eq:polgrad}). Moreover, it is interesting to note that the optimality gap grows with the instance size as solving larger TSP instances is harder.
Additionally, we report the gaps of the best performing policies in Figure \ref{fig:testing}. In the figure, we show the optimality gap of the best solution for 512 test instances over 2,000 steps. Here, results show that we can quickly reduce the optimality gap at the beginning and later steps attempt to fine-tune the best tour. In the experiments, we find the optimal solution for TSP20 instances and stay within optimality gaps of 0.1\% for TSP50 and 0.7\% for TSP100. Overall, our policies can be seen as a solver requiring only random initial solutions and sampling to achieve near-optimal solutions. 

To showcase that, we compare the learned policies with classical 2-opt \textit{First Improvement} (FI) and \textit{Best Improvement} (BI) heuristics, which select the first and best cost-reducing 2-opt operation, respectively. 
Since local search methods can get stuck in local optima, we include a version of the heuristics using \textit{restarts}. That is,  we restart the search at a random solution as soon as we reach a local optimum. We run all heuristics and learned policies on 512 TSP100 instances for a maximum of 1,000 steps starting from the same solutions. The boxplots in Figure \ref{fig:boxplot} depict the results. We observe that our policy (TSP100-Policy) outperforms classical 2-opt heuristics finding tours with lower median and less dispersion. These results support our initial hypothesis that considering future rewards in the choice of 2-opt moves leads to better solutions. Moreover, our method avoids the worst case $O(n^2)$ complexity of selecting the next solution of FI and BI.

\paragraph{Comparison to Classical Heuristics, Exact and Learning Methods}
We report results on the same 10,000 instances for each TSP size as in \cite{kool2018attention} and rerun the optimal results obtained by Concorde to derive optimality gaps. We compare against Nearest, Random and Farthest Insertion constructions heuristics. 
and include the vehicle routing solver of OR-Tools \citep{ortools} containing 2-opt and LKH as improvement heuristics \citep{Bello2017WorkshopT}. 
We add to the comparison recent deep learning methods based on construction and improvement heuristics, including supervised \citep{vinyals2015pointer,joshi2019efficient} and reinforcement \citep{wu2019learning, kool2018attention,deudon2018learning,khalil2017learning,Bello2017WorkshopT} learning methods. We note, however, that supervised learning is not ideal for combinatorial problems due to the lack of optimal labels for large problems.
Previous works to \cite{kool2018attention} are presented with their reported running times and optimality gaps as in the original paper. For recent works, we present the optimality gaps and running times as reported in \citep{kool2018attention,joshi2019efficient, wu2019learning}. We report previous results using greedy, sampling and search decoding and refer to the methods by their neural network architecture. We note that the test dataset used in \cite{wu2019learning} is not the same but the data generation process and size are identical. This fact allied with the high number of samples decreases the variance of the results. We focus our attention on GAT \citep{kool2018attention} and GAT-T \citep{wu2019learning} (GAT-Transformer) representing the best construction and improvement heuristic, respectively.

Our results, in Table \ref{table:comparison},  show that with only 500 steps our method outperforms traditional construction heuristics, learning methods with greedy decoding and OR-Tools achieving $0.01\%$, $0.36\%$ and $1.84\%$ optimality gap for TSP20, TSP50 and TSP100, respectively. Moreover, we outperform GAT-T requiring half the number of steps (500 vs 1,000). We note that with 500 steps, our method also outperforms all previous reinforcement learning methods using sampling or search, including GAT \citep{deudon2018learning} applying 2-opt local search on top of generated tours. Our method only falls short of the supervised learning method GCN \citep{joshi2019efficient}, using beam search and shortest tour heuristic. However, GCN \citep{joshi2019efficient}, similar to samples in GAT \citep{kool2018attention}, uses a beam width of 1,280, i.e. it samples more solutions. Increasing the number of samples (steps) increases the performance of our method. When sampling 1,000 steps (280 samples short of GCN \citep{joshi2019efficient} and GAT \citep{kool2018attention}) we outperform all previous methods that do no employ further local search improvement and perform on par with GAT-T on TSP50, using 5,000 samples (5 times as many samples). For TSP100, sampling 1,000 steps results in a lower optimality gap ($1.26\%$) than all compared methods. Lastly, increasing the sample size to 2,000  results in even lower gaps, $0.00\%$ (TSP20), $0.12\%$ (TSP50)  and $0.87\%$ (TSP100). 

\begin{table}[!ht]
\addtolength{\tabcolsep}{-5pt} 
\centering
\caption{Performance of TSP methods w.r.t. Concorde. 
\emph{Type}: \textbf{SL}: Supervised Learning, \textbf{RL}: Reinforcement Learning, \textbf{S}: Sampling, \textbf{G}: Greedy, \textbf{B}: Beam Search, \textbf{BS}: \textbf{B} and Shortest Tour and \textbf{T}: 2-opt Local Search. \emph{Time}: Time to solve 10,000 instances reported in \citep{kool2018attention,joshi2019efficient,wu2019learning} and ours. 
}
\label{table:comparison}
\begin{adjustbox}{max width=\textwidth}
\begin{tabular}{@{}llc|ccc|ccc|ccc@{}}
\cmidrule[\heavyrulewidth]{2-12}
&\multirow{2}{*}{Method} & \multirow{2}{*}{Type} & \multicolumn{3}{c|}{TSP20} & \multicolumn{3}{c|}{TSP50} & \multicolumn{3}{c}{TSP100} \\
 & & & Cost & Gap & Time & Cost & Gap & Time & Cost & Gap & Time \\
\cmidrule{2-12}
&Concorde \citep{applegate2006traveling} & Solver &  $3.84$ & $0.00 \%$ &(1m)& $5.70$ & $0.00 \%$ &(2m)& $7.76$ & $0.00 \%$ & (3m)  \\
\cmidrule{2-12}
\multirow{4}{*}{\rotatebox[origin=c]{90}{Heuristics}}
&OR-Tools \citep{ortools} & S &  $3.85$ & $0.37 \%$ &  & $5.80$ & $1.83 \%$ &  & $7.99$ & $2.90 \%$ &  \\

&Nearest Insertion & G &  $4.33$ & $12.91 \%$ & (1s) & $6.78$ & $19.03 \%$ & (2s)  & $9.46$ & $21.82 \%$ & (6s)  \\
&Random Insertion & G &  $4.00$ & $4.36 \%$ & (0s) & $6.13$ & $7.65 \%$ & (1s) & $8.52$ & $9.69 \%$ & (3s) \\
&Farthest Insertion & G &  $3.93$ & $2.36 \%$ & (1s) & $6.01$ & $5.53 \%$ & (2s)  & $8.35$ & $7.59 \%$ & (7s) \\
\cmidrule{2-12}
\multirow{6}{*}{\rotatebox[origin=c]{90}{Const.$+$Greedy}}
&PtrNet \citep{vinyals2015pointer} & SL &  $3.88$ & $1.15 \%$ &  & $7.66$ & $34.48 \%$ &  & \multicolumn{3}{c}{-} \\
&GCN \citep{joshi2019efficient}  & SL  &  $3.86$ & $0.60 \%$ & (6s) & $5.87$ & $3.10 \%$ & (55s) & $8.41$ & $8.38 \%$ & (6m) \\
&PtrNet \citep{Bello2017WorkshopT} & RL  &  $3.89$ & $1.42 \%$ &  & $5.95$ & $4.46 \%$ &  & $8.30$ & $6.90 \%$ &  \\
&S2V \citep{khalil2017learning} & RL  &  $3.89$ & $1.42 \%$ &  & $5.99$ & $5.16 \%$ &  & $8.31$ & $7.03 \%$ &  \\
&GAT \citep{deudon2018learning} & RL,T  &  $3.85$ & $0.42 \%$ & (4m) & $5.85$ & $2.77 \%$ & (26m) & $8.17$ & $5.21 \%$ & (3h) \\
&GAT \citep{kool2018attention} & RL  &  $3.85$ & $0.34 \%$ & (0s) & $5.80$ & $1.76 \%$ & (2s) & $8.12$ & $4.53 \%$ & (6s) \\
\cmidrule{2-12}
\multirow{6}{*}{\rotatebox[origin=c]{90}{Const.$+$Search}}
&GCN \citep{joshi2019efficient} & SL,B &  $3.84$ & $0.10 \%$ & (20s) & $5.71$ & $0.26 \%$ & (2m) & $7.92$ & $2.11 \%$ & (10m) \\
&GCN \citep{joshi2019efficient} & SL,BS &  $3.84$ & $0.01 \%$ & (12m) & $\mathbf{5.70}$ & $\mathbf{0.01} \%$ & (18m) & $7.87$ & $1.39 \%$ & (40m) \\
&PtrNet \citep{Bello2017WorkshopT} & RL,S & \multicolumn{3}{c|}{-} & $5.75$ & $0.95 \%$ &  & $8.00$ & $3.03 \%$ &  \\
&GAT \citep{deudon2018learning} & RL,S &  $3.84$ & $0.11 \%$ & (5m) & $5.77$ & $1.28 \%$ & (17m) & $8.75$ & $12.70 \%$ & (56m) \\
&GAT \citep{deudon2018learning} & RL,S,T  &  $3.84$ & $0.09 \%$ & (6m) & $5.75$ & $1.00 \%$ & (32m) & $8.12$ & $4.64 \%$ & (5h) \\
&GAT \{1280\} \citep{kool2018attention} & RL,S  &  $3.84$ & $0.08 \%$ & (5m) & $5.73$ & $0.52 \%$ & (24m) & $7.94$ & $2.26 \%$ & (1h) \\
\cmidrule{2-12}
\multirow{6}{*}{\rotatebox{90}{Impr.$+$Sampling}}
&GAT-T \{1000\} \citep{wu2019learning} & RL &  $3.84$ & $0.03 \%$ &  (12m) & $5.75$ & $0.83 \%$ & (16m) & $8.01$ & $3.24 \%$ & (25m) \\
&GAT-T \{3000\} \citep{wu2019learning} & RL &  $3.84$ & $0.00 \%$ &  (39m) & $5.72$ & $0.34 \%$ & (45m) & $7.91$ & $1.85 \%$ & (1h) \\
&GAT-T \{5000\} \citep{wu2019learning} & RL &  $3.84$ & $0.00 \%$ &  (1h) & $5.71$ & $0.20 \%$ & (1h) & $7.87$ & $1.42 \%$ & (2h)  \\
\cmidrule{2-12}
&Ours \{500\} & RL & $3.84$ & $0.01 \%$ & (5m)& $5.72$ & $0.36 \%$ & (7m)  & $7.91$ & $1.84 \%$ & (10m)\\

&Ours \{1000\}  & RL &  $\mathbf{3.84}$ & $\mathbf{0.00 \%}$ & (10m) & $5.71$ & $0.21 \%$ & (13m)  & $7.86$  & $1.26\%$  &  (21m)\\
&Ours \{2000\} & RL&  $\mathbf{3.84}$ & $\mathbf{0.00} \%$ & (15m) & $5.70$ & $0.12 \%$ & (29m)& $\mathbf{7.83}$ & $\mathbf{0.87\%}$  &  (41m)\\
\cmidrule[\heavyrulewidth]{2-12}
\end{tabular}
\end{adjustbox}
\end{table}


\paragraph{Testing Learned Policies on Larger Instances}
Since we are interested in learning general policies that can solve the TSP regardless of its size, we test the performance of our policies when learning on TSP50 instances (TSP50-Policy) and applying on larger TSP100 instances. Results, in Table \ref{table:train50}, show that we can extract general enough information to still perform well on 100 nodes. Similar to a TSP100-Policy, our TSP50-Policy can outperform previous reinforcement learning construction approaches and requires fewer samples. With 1,000 samples TSP50-Policy performs similarly to GAT-T \citep{wu2019learning} using 3,000 samples, at $1.86\%$ optimality gap. These results are closer to optimal than previous learning methods without further local search improvement as in GCN \citep{joshi2019efficient}. When increasing to 2,000 steps, we outperform all compared methods at $1.37\%$ optimality gap.


\begin{table}[!htb]
\begin{minipage}{.48\linewidth}
\centering
\caption{Performance of policies trained on 50 and 100 nodes on TSP100 instances.}
\label{table:train50}
\begin{tabular}{lccccc}
\toprule
& \multicolumn{2}{c}{TSP100-Policy} & &\multicolumn{2}{c}{TSP50-Policy} \\
\cmidrule{2-3} \cmidrule{5-6}
Steps & Cost & Gap && Cost & Gap \\
\midrule
500 &  $7.91$ & $1.84 \%$ & &  $7.98$ &  $2.78 \%$  \\ 
1000 & $7.86$  & $1.26\%$ & & $7.91$ &  $1.86 \%$ \\ 
2000 & $7.83$ & $0.87 \%$ & & $7.87$ &  $1.37\%$  \\ 
\bottomrule
\end{tabular}
\end{minipage}\hfill
\begin{minipage}{.48\linewidth}
\addtolength{\tabcolsep}{-3pt} 
\centering
\caption{Ablation studies on 512 TSP50 instances running policies for 1,000 steps.}
	\label{tab:ablation} 
\begin{adjustbox}{max width=\textwidth}
	\begin{tabular}{lccccc} 
		
		\toprule 
		& \multicolumn{2}{c}{Epoch: 10} && \multicolumn{2}{c}{Epoch: 200} \\
        \cmidrule{2-3} \cmidrule{5-6}
					&	Opt. Gap (\%)	&	Cost && Opt. Gap (\%)	&	Cost	\\
		\midrule
		Proposed 	& \bf 3.00 $\pm$ 0.08 	& \bf 5.87	&& \bf	0.22 $\pm$ 0.01& \bf 5.72 \\
		\midrule
		(a) w/o bi-LSTM		& 203.87 $\pm$ 0.61  & 17.33	&&  134.42 $\pm$ 0.56	& 13.37 \\
		(b) w/o GCN		& 9.74 $\pm$ 0.08	&  6.26	&&  0.30 $\pm$ 0.01	& 5.72 \\
		(c) w/o bidirectional		& 17.94 $\pm$ 0.15  &  6.73	&&   2.20 $\pm$ 0.05	& 5.82 \\
		(d) w/o best solution		& 4.55 $\pm$ 0.04	&  5.96	&&  \bf 0.22 $\pm$ 0.02	&  \bf 5.72 \\
		(e) shared encoder		& 5.15 $\pm$ 0.06 & 6.00  &&  0.23 $\pm$ 0.01	&  5.72 \\
		
		
		\bottomrule
	\end{tabular}
\end{adjustbox}
\end{minipage}
\end{table}

\paragraph{Running Times and Sample Efficiency}
Comparing running times is difficult due to varying hardware and implementations among different approaches. In Table \ref{table:comparison}, we report the running times to solve 10,000 instances as reported in \citep{kool2018attention,joshi2019efficient,wu2019learning} and ours. We focus on learning methods, as classical heuristics and solvers are efficiently implemented using multi-threaded CPUs. 
We note that our method cannot compete in speed with greedy methods as we start from poor solutions and require sampling to find improved solutions. This is neither surprising nor discouraging, as one can see these methods as a way to generate initial solutions for an improvement heuristic like ours.
We note, however, that while sampling 1,000 steps, our method is faster than GAT-T \citep{wu2019learning} even though we use a less powerful GPU (RTX 2080Ti vs Tesla V100). Moreover, our method requires fewer samples to achieve superior performance. The comparison to GAT \citep{kool2018attention} is not so straightforward as they use a GTX 1080Ti and different number of samples. For this reason, we run GAT \citep{kool2018attention} using our hardware and report running times sampling the same number of solutions in Table \ref{table:kool}.
Our method is slower for TSP20 and TSP50 sampling 2,000 solutions. However, as we reach TSP100, our method can be computed faster and, overall, requires less time to produce shorter tours. 

\begin{table}[!htb]
\begin{minipage}{.48\linewidth}
\centering
\caption{Performance of GAT \citep{kool2018attention} vs our method. Results are compared on the same hardware sampling the same number of solutions.}
\label{table:kool}
\begin{adjustbox}{max width=\textwidth}
\begin{tabular}{l|cc|cc|cc}
\toprule
\multirow{2}{*}{Method} & \multicolumn{2}{c|}{TSP20} & \multicolumn{2}{c|}{TSP50} & \multicolumn{2}{c}{TSP100} \\
 & Cost & Time & Cost & Time & Cost & Time \\
\midrule
GAT \{500\} &  $3.839$ & \textbf{(3m) }& $5.727$ & (10m) & $7.955$ & (27m) \\
Ours \{500\} & $3.836$ & (5m) & $5.716$ & \textbf{(7m)} &$7.907$ & \textbf{(10m)}\\
\midrule
GAT \{1,000\} &  $3.838$ & \textbf{(4m) }& $5.725$ & (14m) & $7.947$ & (42m) \\
Ours \{1,000\} & $3.836$ & (10m) & $5.708$ & \textbf{(13m) }& $7.861$& \textbf{(21m)}\\
\midrule
GAT \{2,000\}  &  $3.838$ &\textbf{ (5m) }& $5.722$ & \textbf{(22m)} & $7.939$ & (1h13m) \\
Ours \{2,000\} & $3.836$ & (15m) & $5.703$& (29m) & $7.832$& \textbf{(41m)}\\
\bottomrule
\end{tabular}
\end{adjustbox}

\end{minipage}\hfill
\begin{minipage}{.48\linewidth}
\addtolength{\tabcolsep}{-3pt} 
\centering
\caption{Performance of OR-Tools vs our method on TSPLib. See footnote \ref{fn_tsplib}.}
\label{table:tsplib}
\begin{adjustbox}{max width=\textwidth}
\begin{tabular}{l|c|cc}
\toprule
\multicolumn{1}{c|}{Instance} & Opt. & \begin{tabular}[c]{@{}c@{}}Ours \{2000\}\end{tabular} & \begin{tabular}[c]{@{}c@{}}OR-Tools\end{tabular} \\ \midrule
eil51         & 426     & \textbf{427}              & 439    \\
berlin52      & 7,542   & 7,974            & \textbf{7,944}   \\
pr76          & 108,159 & 111,085          & \textbf{110,948} \\
rd100         & 7,910   & \textbf{7,944}   & 8,221            \\
eil101        & 629     & \textbf{635}     & 650              \\
lin105        & 14,379  & 16,156  & \textbf{15,363}           \\
ch130         & 6,110   & \textbf{6,175}   & 6,329            \\
pr144         & 58,537  & 61,207  & \textbf{59,286}           \\
ts225         & 126,643 & \textbf{127,731} & 127,763          \\
a280          & 2,579   & 2,898   & \textbf{2,742}            \\
\midrule
Avg. Opt. Gap & 0.00\%     & 4.56\%           & 3.79\%          \\ \bottomrule
\end{tabular}%
\end{adjustbox}
\end{minipage}
\end{table}

\paragraph{Ablation Study} In Table \ref{tab:ablation}, we present an ablation study of the proposed method. We measure the performance at the beginning and towards the end of training, i.e. at epochs 10 and 200, rolling out policies for 1,000 steps for 512 TSP50 instances and 10 trials. We observe that removing the LSTM (a) affects performance the most leading to a large 134.42\% gap at epoch 200. Removing the GCN component (b) has a lower influence but also reduces the overall quality of policies, reaching 0.30\% optimality gap. We then test the effect of the bidirectional LSTM (c) replacing it by a single LSTM. In this case, gaps are even higher, at 2.20\%, suggesting that encoding the symmetry of the tours is important. We also compare to two variants of the proposed model, one that does not take as input the best solution (d) and one that shares the parameters of the encoding units (e). For these cases, we note that the final performance is similar to the proposed method, i.e. 0.22\% optimality gap. However, in our experiments, the proposed method can achieve better policies faster, reaching a 3.0\% gap at epoch 10, whereas (d) and (e) yield policies at the 4.55\% and 5.15\% level, respectively.

\paragraph{Generalization to Real-world TSP instances} In Table \ref{table:tsplib}, we study the performance of our method on TSPlib \citep{reinelt1991tsplib} instances. In general, these instances come from different node distributions than those seen during training and it is unclear whether our learned policies can be reused for these cases. We compare the results of the policy trained on TSP100 sampling actions for 2,000 steps to results obtained from OR-Tools. We note that for the 10 instances tested, our method outperforms OR-Tools in 5 instances. These results are encouraging as OR-Tools is a very specialized heuristic solver. When we compare optimality gaps (4.56\% vs 3.79\%)\footnote{We perform a more extensive comparison using 35 TSPlib instances in the Supplementary Materials. On the 35 instances the gaps are 8.61\% (ours) and 3.70\% (OR-Tools).\label{fn_tsplib}} we see that our learned policies are not too far from OR-Tools even though our method never trains on instances with more than 100 nodes. The difference in performance increases for large instances, indicating that fine-tuning or training policies for more nodes and different distributions can potentially reduce this difference. However, similar to results in Table \ref{table:train50}, our method still can achieve good results on instances with more than 100 nodes, such as ts225 (0.86\% gap).

\section{Conclusions and Future Work}

In this work, we introduced a novel deep reinforcement learning approach for approximating a 2-opt improvement heuristic for the Euclidean Traveling Salesman Problem (TSP). 
We proposed a neural architecture with graph and sequence embeddings capable of outperforming state-of-the-art learned construction and improvement heuristics requiring fewer samples. Our learned heuristics also outperform classical 2-opt ones reaching lower optimality gaps.

Expanding the proposed neural architecture to sample $k$-opt operations is an interesting topic for future work. Moreover, exploring general improvement heuristics that can be applied to a large number of combinatorial problems is another interesting idea for further development. One drawback of our policy gradient method is the large number of samples required to train a good policy. As a future direction, we intend to explore methods that can be more sample efficient and can learn good policies requiring less training time. Lastly, we point out that future work on learning heuristics can be useful when solving problems where standard solvers are not performant, e.g., a TSP with on-route stochastic travel costs. 

%




\bibliography{acml20}

\begin{thebibliography}{27}
\providecommand{\natexlab}[1]{#1}
\providecommand{\url}[1]{\texttt{#1}}
\expandafter\ifx\csname urlstyle\endcsname\relax
  \providecommand{\doi}[1]{doi: #1}\else
  \providecommand{\doi}{doi: \begingroup \urlstyle{rm}\Url}\fi

\bibitem[Angeniol et~al.(1988)Angeniol, Vaubois, and
  Le~Texier]{angeniol1988self}
Bernard Angeniol, Gael De La~Croix Vaubois, and Jean-Yves Le~Texier.
\newblock Self-organizing feature maps and the travelling salesman problem.
\newblock \emph{Neural Networks}, 1\penalty0 (4):\penalty0 289--293, 1988.

\bibitem[Applegate et~al.(2006)Applegate, Bixby, Chvatal, and
  Cook]{applegate2006traveling}
David~L Applegate, Robert~E Bixby, Vasek Chvatal, and William~J Cook.
\newblock \emph{The traveling salesman problem: a computational study}.
\newblock Princeton university press, 2006.

\bibitem[Arora(1998)]{arora1998polynomial}
Sanjeev Arora.
\newblock Polynomial time approximation schemes for euclidean traveling
  salesman and other geometric problems.
\newblock \emph{Journal of the ACM (JACM)}, 45\penalty0 (5):\penalty0 753--782,
  1998.

\bibitem[Bello and Pham(2017)]{Bello2017WorkshopT}
Irwan Bello and Hieu Pham.
\newblock Neural combinatorial optimization with reinforcement learning.
\newblock In \emph{Proceedings of the 5th International Conference on Learning
  Representations (ICLR)}, 2017.

\bibitem[Bengio et~al.(2018)Bengio, Lodi, and Prouvost]{bengio2018machine}
Yoshua Bengio, Andrea Lodi, and Antoine Prouvost.
\newblock Machine learning for combinatorial optimization: a methodological
  tour d'horizon.
\newblock \emph{arXiv preprint arXiv:1811.06128}, 2018.

\bibitem[Deudon et~al.(2018)Deudon, Cournut, Lacoste, Adulyasak, and
  Rousseau]{deudon2018learning}
Michel Deudon, Pierre Cournut, Alexandre Lacoste, Yossiri Adulyasak, and
  Louis-Martin Rousseau.
\newblock Learning heuristics for the tsp by policy gradient.
\newblock In \emph{Proceedings of the 15th International Conference on the
  Integration of Constraint Programming, Artificial Intelligence, and
  Operations Research (CPAIOR)}, pages 170--181, 2018.

\bibitem[Hansen and Mladenovi{\'c}(2006)]{hansen2006first}
Pierre Hansen and Nenad Mladenovi{\'c}.
\newblock First vs. best improvement: An empirical study.
\newblock \emph{Discrete Applied Mathematics}, 154\penalty0 (5):\penalty0
  802--817, 2006.

\bibitem[Helsgaun(2009)]{helsgaun2009general}
Keld Helsgaun.
\newblock General k-opt submoves for the lin--kernighan tsp heuristic.
\newblock \emph{Mathematical Programming Computation}, 1\penalty0
  (2-3):\penalty0 119--163, 2009.

\bibitem[Hochreiter and Schmidhuber(1997)]{hochreiter1997long}
Sepp Hochreiter and J{\"u}rgen Schmidhuber.
\newblock Long short-term memory.
\newblock \emph{Neural computation}, 9\penalty0 (8):\penalty0 1735--1780, 1997.

\bibitem[Hopfield and Tank(1985)]{hopfield1985neural}
John~J Hopfield and David~W Tank.
\newblock Neural computation of decisions in optimization problems.
\newblock \emph{Biological cybernetics}, 52\penalty0 (3):\penalty0 141--152,
  1985.

\bibitem[Jia et~al.(2018)Jia, Song, He, Wang, Rong, Zhou, Xie, Guo, Yang, Yu,
  et~al.]{jia2018highly}
Xianyan Jia, Shutao Song, Wei He, Yangzihao Wang, Haidong Rong, Feihu Zhou,
  Liqiang Xie, Zhenyu Guo, Yuanzhou Yang, Liwei Yu, et~al.
\newblock Highly scalable deep learning training system with mixed-precision:
  Training imagenet in four minutes.
\newblock \emph{arXiv preprint arXiv:1807.11205}, 2018.

\bibitem[Joshi et~al.(2019)Joshi, Laurent, and Bresson]{joshi2019efficient}
Chaitanya~K Joshi, Thomas Laurent, and Xavier Bresson.
\newblock An efficient graph convolutional network technique for the travelling
  salesman problem.
\newblock \emph{arXiv:1906.01227}, 2019.

\bibitem[Khalil et~al.(2017)Khalil, Dai, Zhang, Dilkina, and
  Song]{khalil2017learning}
Elias Khalil, Hanjun Dai, Yuyu Zhang, Bistra Dilkina, and Le~Song.
\newblock Learning combinatorial optimization algorithms over graphs.
\newblock In \emph{Proceedings of the 31st Conference on Neural Information
  Processing Systems (NIPS)}, pages 6348--6358, 2017.

\bibitem[Kingma and Ba(2015)]{kingma2015adam}
Diederik~P Kingma and Jimmy Ba.
\newblock Adam: A method for stochastic optimization.
\newblock In \emph{International Conference on Machine Learning}, 2015.

\bibitem[Kipf and Welling(2017)]{kipf2016semi}
Thomas~N. Kipf and Max Welling.
\newblock Semi-supervised classification with graph convolutional networks.
\newblock In \emph{Proceedings of the 5th International Conference on Learning
  Representations, {(ICLR)}}, 2017.

\bibitem[Kool et~al.(2019)Kool, van Hoof, and Welling]{kool2018attention}
Wouter Kool, Herke van Hoof, and Max Welling.
\newblock Attention, learn to solve routing problems!
\newblock In \emph{Proceedings of the 7th International Conference on Learning
  Representations (ICLR)}, 2019.

\bibitem[La~Maire and Mladenov(2012)]{la2012comparison}
Bert~FJ La~Maire and Valeri~M Mladenov.
\newblock Comparison of neural networks for solving the travelling salesman
  problem.
\newblock In \emph{11th Symposium on Neural Network Applications in Electrical
  Engineering}, pages 21--24. IEEE, 2012.

\bibitem[Lin and Kernighan(1973)]{lin1973effective}
Shen Lin and Brian~W Kernighan.
\newblock An effective heuristic algorithm for the traveling-salesman problem.
\newblock \emph{Operations research}, 21\penalty0 (2):\penalty0 498--516, 1973.

\bibitem[Lombardi and Milano(2018)]{lombardi2018boosting}
Michele Lombardi and Michela Milano.
\newblock Boosting combinatorial problem modeling with machine learning.
\newblock In \emph{Proceedings of the 27th International Joint Conference on
  Artificial Intelligence (IJCAI)}, pages 5472--5478, 2018.

\bibitem[Mnih et~al.(2016)Mnih, Badia, Mirza, Graves, Lillicrap, Harley,
  Silver, and Kavukcuoglu]{mnih2016asynchronous}
Volodymyr Mnih, Adria~Puigdomenech Badia, Mehdi Mirza, Alex Graves, Timothy
  Lillicrap, Tim Harley, David Silver, and Koray Kavukcuoglu.
\newblock Asynchronous methods for deep reinforcement learning.
\newblock In \emph{Proceedings of the 33rd International Conference on Machine
  Learning (ICML)}, pages 1928--1937, 2016.

\bibitem[Papadimitriou(1977)]{papadimitriou1977euclidean}
Christos~H Papadimitriou.
\newblock The euclidean travelling salesman problem is np-complete.
\newblock \emph{Theoretical Computer Science}, 4\penalty0 (3):\penalty0
  237--244, 1977.

\bibitem[Perron and Furnon()]{ortools}
Laurent Perron and Vincent Furnon.
\newblock Or-tools.
\newblock URL \url{https://developers.google.com/optimization/}.

\bibitem[Reinelt(1991)]{reinelt1991tsplib}
Gerhard Reinelt.
\newblock Tsplib—a traveling salesman problem library.
\newblock \emph{ORSA journal on computing}, 3\penalty0 (4):\penalty0 376--384,
  1991.

\bibitem[Veli{\v{c}}kovi{\'{c}} et~al.(2018)Veli{\v{c}}kovi{\'{c}}, Cucurull,
  Casanova, Romero, Li{\`{o}}, and Bengio]{velickovic2018graph}
Petar Veli{\v{c}}kovi{\'{c}}, Guillem Cucurull, Arantxa Casanova, Adriana
  Romero, Pietro Li{\`{o}}, and Yoshua Bengio.
\newblock {Graph Attention Networks}.
\newblock In \emph{Proceedings of the 6th International Conference on Learning
  Representations (ICLR)}, 2018.

\bibitem[Vinyals et~al.(2015)Vinyals, Fortunato, and
  Jaitly]{vinyals2015pointer}
Oriol Vinyals, Meire Fortunato, and Navdeep Jaitly.
\newblock Pointer networks.
\newblock In \emph{Proceedings of the 29th Conference on Neural Information
  Processing Systems (NIPS)}, pages 2692--2700, 2015.

\bibitem[Williams(1992)]{williams1992simple}
Ronald~J Williams.
\newblock Simple statistical gradient-following algorithms for connectionist
  reinforcement learning.
\newblock \emph{Machine learning}, 8\penalty0 (3-4):\penalty0 229--256, 1992.

\bibitem[Wu et~al.(2019)Wu, Song, Cao, Zhang, and Lim]{wu2019learning}
Yaoxin Wu, Wen Song, Zhiguang Cao, Jie Zhang, and Andrew Lim.
\newblock Learning improvement heuristics for solving the travelling salesman
  problem.
\newblock \emph{arXiv:1912.05784}, 2019.

\end{thebibliography}

\section*{Supplementary Material}
\label{sec:supplement}

\subsection*{Performance on TSPlib Instances}

We show the performance of our method in comparison to OR-Tools \citep{ortools} on 35 TSPlib \citep{reinelt1991tsplib} instances in Table \ref{tab:sup} below. 

\begin{table}[H]

\centering
\caption{Performance of OR-Tools vs our method on TSPlib instances.}
\resizebox{0.4\textwidth}{!}{%

\begin{tabular}{l|c|cc}
\toprule
Instance      & Opt.     & Ours \{2,000\}   & OR-Tools          \\ \midrule
eil51         & 426     & \textbf{427}     & 439              \\
berlin52      & 7,542   & 7,974            & \textbf{7,944}   \\
st70          & 675     & \textbf{680}     & 683              \\
eil76         & 538     & 552              & \textbf{548}     \\
pr76          & 108,159 & 111,085          & \textbf{110,948} \\
rat99         & 1,211   & 1,388            & \textbf{1,284}   \\
rd100         & 7,910   & \textbf{7,944}   & 8,221            \\
kroA100       & 21,282  & 23,751           & \textbf{21,960}  \\
kroB100       & 22,141  & 23,790           & \textbf{22,945}  \\
kroC100       & 20,749  & 22,672           & \textbf{21,699}  \\
kroD100       & 21,294  & 23,334           & \textbf{22,439}  \\
kroE100       & 22,068  & 23,253           & \textbf{22,551}  \\
eil101        & 629     & \textbf{635}     & 650              \\
lin105        & 14,379  & 16,156           & \textbf{15,363}  \\
pr107         & 44,303  & 54,378           & \textbf{44,573}  \\
pr124         & 59,030  & 59,516           & \textbf{60,413}  \\
bier127       & 118,282 & \textbf{121,122} & 121,729          \\
ch130         & 6,110   & \textbf{6,175}   & 6,329            \\
pr136         & 96,772  & \textbf{98,453}  & 102,813          \\
pr144         & 58,537  & 61,207           & \textbf{59,286}  \\
ch150         & 6,528   & \textbf{6,597}   & 6,733            \\
kroA150       & 26,524  & 30,078           & \textbf{27,503}  \\
kroB150       & 26,130  & 28,169           & \textbf{26,671}  \\
pr152         & 73,682  & \textbf{75,301}  & 75,832           \\
u159          & 42,080  & \textbf{42,716}  & 43,403           \\
rat195        & 2,323   & 2,955            & \textbf{2,375}   \\
kroA200       & 29,368  & 32,522           & \textbf{29,874}  \\
ts225         & 126,643 & \textbf{127,731} & 127,763          \\
tsp225        & 3,919   & 4,354            & \textbf{4,117}   \\
pr226         & 80,369  & 91,560           & \textbf{83,113}  \\
gil262        & 2,378   & \textbf{2,490}   & 2,517            \\
pr264         & 49,135  & 59,109           & \textbf{51,495}  \\
a280          & 2,579   & 2,898            & \textbf{2,742}   \\
pr299         & 48,191  & 59,422           & \textbf{50,617}  \\
pr439         & 107,217 & 143,590          & \textbf{117,171} \\
\midrule
Avg. Opt. Gap & 0.00\%  & 8.61\%           & 3.70\%           \\ \bottomrule
\end{tabular}%
\label{tab:sup}
}
\end{table}

\end{document}